\documentclass[11pt,a4paper]{article}
\usepackage[hyperref]{acl2020}
\usepackage{times}
\usepackage{latexsym}

\usepackage{microtype}
\usepackage[export]{adjustbox}
\usepackage{booktabs}
\usepackage{titling}
\thanksmarkseries{arabic}
\newcommand*{\affaddr}[1]{#1} 
\newcommand*{\affmark}[1][*]{\textsuperscript{#1}}

\aclfinalcopy 

\title{\textbf{Template-Based Question Generation from Retrieved Sentences for Improved Unsupervised Question Answering}}

\author{\textbf{Alexander R. Fabbri}\affmark{$\dagger$} \thanks{Equal contribution}\protect\phantom{\footnotesize 1}\textsuperscript{,}\thanks{Work done during internship at the AWS AI Labs}
  \quad \textbf{Patrick Ng}\affmark{$\ddagger$}\footnotemark[1]\\
  \quad \textbf{Zhiguo Wang}\affmark{$\ddagger$}
  \quad \textbf{Ramesh Nallapati}\affmark{$\ddagger$}
  \quad \textbf{Bing Xiang}\affmark{$\ddagger$}\\
  \affaddr{\affmark[$\dagger$]Yale University} 
  \affaddr{\affmark[$\ddagger$]AWS AI Labs} \\
  alexander.fabbri@yale.edu, \{patricng,\,zhiguow, rnallapa, bxiang\}@amazon.com
  }

\date{}
\begin{document}
\maketitle
\begin{abstract}
Question Answering (QA) is in increasing demand as the amount of information available online and the desire for quick access to this content grows. A common approach to QA has been to fine-tune a pretrained language model on a task-specific labeled dataset. This paradigm, however, relies on scarce, and costly to obtain, large-scale human-labeled data. We propose an unsupervised approach to training QA models with generated pseudo-training data. We show that generating questions for QA training by applying a simple template on a related, retrieved sentence rather than the original context sentence improves downstream QA performance by allowing the model to learn more complex context-question relationships. Training a QA model on this data gives a relative improvement over a previous unsupervised model in F1 score on the SQuAD dataset by about 14\%, and 20\% when the answer is a named entity, achieving state-of-the-art performance on SQuAD for unsupervised QA.


\end{abstract}
\section{Introduction}
Question Answering aims to answer a question based on a given knowledge source. Recent advances have driven the performance of QA systems to above or near-human performance on QA datasets such as SQuAD \cite{rajpurkar-etal-2016-squad} and Natural Questions \cite{kwiatkowski-etal-2019-natural} thanks to pretrained language models such as BERT \cite{devlin-etal-2019-bert}, XLNet \cite{yang2019xlnet} and RoBERTa \cite{liu2019roberta}. Fine-tuning these language models, however, requires large-scale data for fine-tuning. Creating a dataset for every new domain is extremely costly and practically infeasible. The ability to apply QA models on out-of-domain data in an efficient manner is thus very desirable. This problem may be approached with domain adaptation or transfer learning techniques \cite{chung-etal-2018-supervised} as well as data augmentation \cite{yang-etal-2017-semi,dhingra-etal-2018-simple,wang-etal-2018-multi,alberti-etal-2019-synthetic}.  However, here we expand upon the recently introduced task of unsupervised question answering \cite{lewis-etal-2019-unsupervised} to examine the extent to which synthetic training data alone can be used to train a QA model. 
\begin{figure}[t]
\centering
\includegraphics[width=\columnwidth]{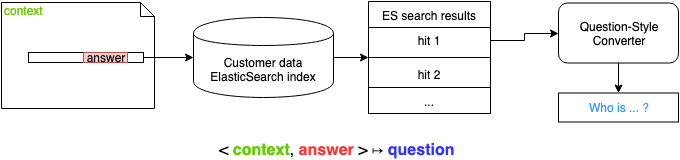}
\caption{Question Generation Pipeline: the original context sentence containing a given answer is used as a query to retrieve a related sentence containing matching entities, which is input into our question-style converter to create QA training data.}
\label{fig:pipeline}
\end{figure}

In particular, we focus on the machine reading comprehension setting in which the context is a given paragraph, and the QA model can only access this paragraph to answer a question. Furthermore, we work on extractive QA, where the answer is assumed to be a contiguous sub-string of the context. A training instance for supervised reading comprehension consists of three components: a \emph{question}, a \emph{context}, and an \emph{answer}. For a given dataset domain, a collection of documents can usually be easily obtained, providing \emph{context} in the form of paragraphs or sets of sentences. \emph{Answers} can be gathered from keywords and phrases from the context. We focus mainly on factoid QA; the question concerns a concise fact. In particular, we emphasize questions whose answers are named entities, the majority type of factoid questions. Entities can be extracted from text using named entity recognition (NER) techniques as the training instance's \emph{answer}.  Thus, the main challenge, and the focus of this paper, is creating a relevant \emph{question} from a \emph{(context, answer)} pair in an unsupervised manner.

\par
Recent work of \cite{lewis-etal-2019-unsupervised} uses style transfer for generating questions for \emph{(context, answer)} pairs but shows little improvement over applying a much simpler question generator which drops, permutates and masks words. We improve upon this paper by proposing a simple, intuitive, \textbf{retrieval and template-based question generation} approach, illustrated in Figure \ref{fig:pipeline}. The idea is to retrieve a sentence from the corpus similar to the current context, and then generate a \emph{question} based on that sentence. Having created a \emph{question} for all \emph{(context, answer)} pairs, we then fine-tune a pretrained BERT model on this data and evaluate on the SQuAD v1.1 dataset \cite{rajpurkar-etal-2016-squad}.


\par
Our contributions are as follows: we introduce a retrieval, template-based framework which achieves state-of-the-art results on SQuAD for unsupervised models, particularly when the answer is a named entity. We perform ablation studies to determine the effect of components in template question generation. We are releasing our synthetic training data and code.\footnote{\url{https://github.com/awslabs/unsupervised-qa}}
\section{Unsupervised QA Approach}
We focus on creating high-quality, non-trivial questions which will allow the model to learn to extract the proper answer from a context-question pair. 
\par
\textbf{Sentence Retrieval:}
A standard cloze question can be obtained by taking the original sentence in which the answer appears from the context and masking the answer with a chosen token. However, a model trained on this data will only learn text matching and how to fill-in-the-blank, with little generalizability. For this reason, we chose to use a retrieval-based approach to obtain a sentence similar to that which contains the answer, upon which to create a given question. For our experiments, we focused on answers which are named entities, which has proven to be a useful prior assumption for downstream QA performance \cite{lewis-etal-2019-unsupervised} confirmed by our initial experiments. 
First, we indexed all of the sentences from a Wikipedia dump using the ElasticSearch search engine. We also extract named entities for each sentence in both the Wikipedia corpus and the sentences used as queries. We assume access to a named-entity recognition system, and in this work make use of the spaCy\footnote{\url{https://spacy.io}} NER pipeline. Then, for a given context-answer pair, we query the index, using the original context sentence as a query, to return a sentence which (1) contains the answer, (2) does not come from the \emph{context}, and (3) has a lower than 95\% F1 score with the query sentence to discard highly similar or plagiarized sentences. Besides ensuring that the retrieved sentence and query sentence share the answer entity, we require that at least one additional matching entity appears in both the query sentence and in the entire context, and we perform ablation studies on the effect of this matching below. These retrieved sentences are then fed into our question-generation module. 
\begin{figure}[t!]
\centering
\includegraphics[width=\columnwidth]{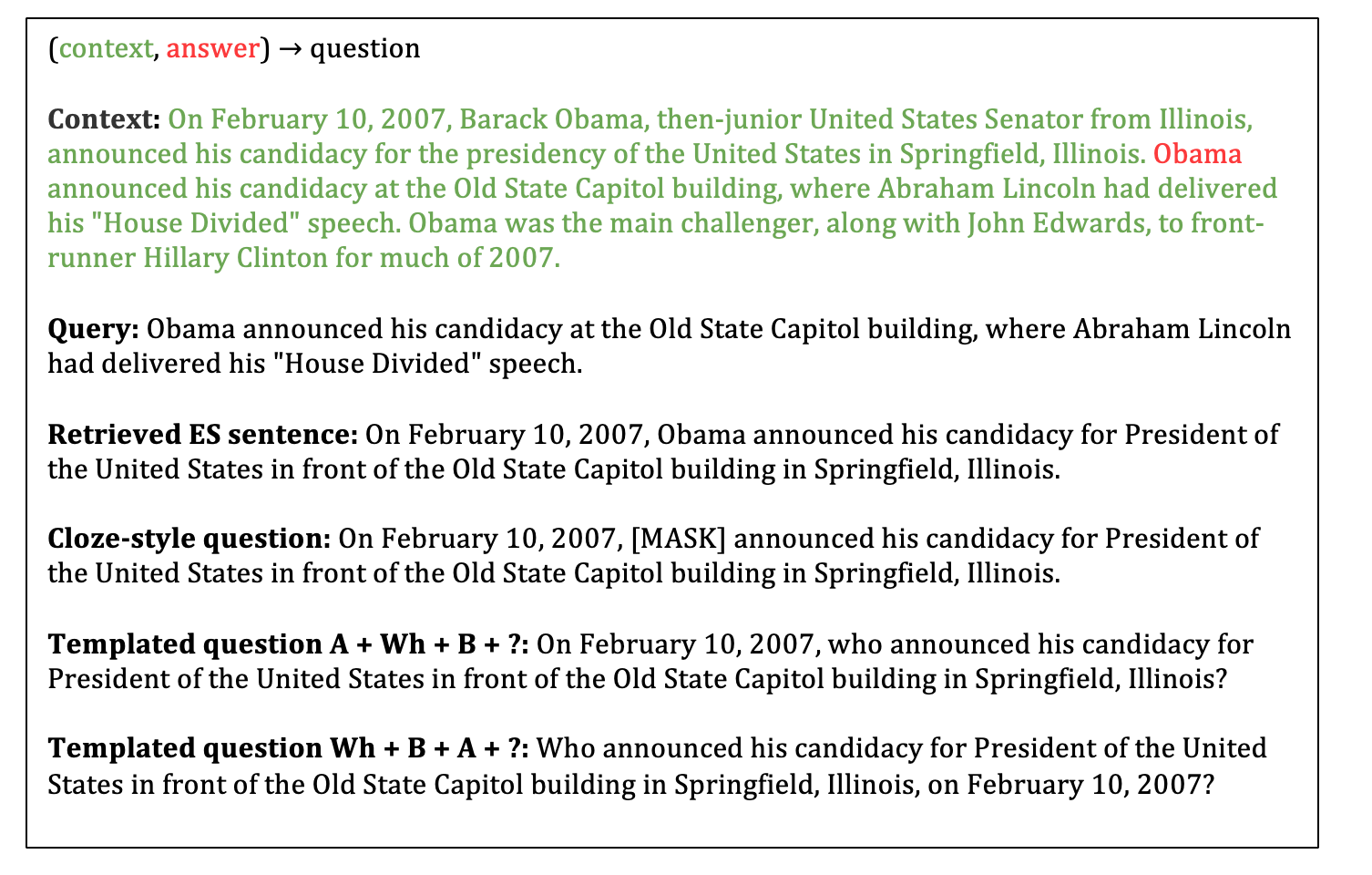}
\caption{Example of synthetically generated questions using generic cloze-style questions as well as a template-based approach.}
\label{fig:question-example}
\end{figure}
\par
\textbf{Template-based Question Generation:}
 We consider several question styles (1) generic cloze-style questions where the answer is replaced by the token ``[MASK]", (2)
templated question ``Wh+B+A+?" as well as variations on the ordering of this template, as shown in Figure \ref{fig:question-example}. Given the retrieved sentence in the form of
\texttt{{[}Fragment\ A{]}\ {[}Answer{]}\ {[}Fragment\ B{]}}, the templated question ``Wh+B+A+?" replaces the \emph{answer} with a Wh-component (e.g., what, who, where), which depends on the entity type of the \emph{answer} and places the Wh-component at the beginning of the question, followed by sentence \texttt{Fragment\ B} and \texttt{Fragment\ A}. For the choice of wh-component, we sample a bi-gram based on prior probabilities of that bi-gram being associated with the named-entity type of the answer. This prior probability is calculated based on named-entity and question bi-gram starters from the SQuAD dataset. This information does not make use of the full context-question-answer and can be viewed as prior information, not disturbing the integrity of our unsupervised approach. Additionally, the choice of wh component does not significantly affect results. For template-based approaches, we also experimented with clause-based templates but did not find significant differences in performance. 
\section{Experiments}
\par
\textbf{Settings:} For all downstream question answering models, we fine-tune a pretrained BERT model using the Transformers repository \cite{Wolf2019HuggingFacesTS}  and report ablation study numbers using the base-uncased version of BERT, consistent with \cite{lewis-etal-2019-unsupervised}. All models are \emph{trained and validated} on \emph{generated pairs of questions and answers} along with their contexts \emph{tested} on the \emph{SQuAD development set}. The training set differs for each ablation study and will be described below, while the validation dataset is a random set of 1,000 template-based generated data points, which is consistent across all ablation studies. We train all QA models for 2 epochs, checkpointing the models every 500 steps and choosing the checkpoint with the highest F1 score on the validation set as the best model. All ablation studies are averaged over two training runs with different seeds. Unless otherwise stated, experiments are performed using 50,000 synthetic QA training examples, as initial models performed best with this amount. We will make this generated training data public. 
\subsection{Model Analysis}
\textbf{Effect of retrieved sentences:} We test the effect of retrieved vs original sentences as input to question generation when using generic cloze questions. As shown in Table \ref{tab:results_cloze_orig_ret}, using retrieved sentences improves over using the original sentence, reinforcing our motivation that a retrieved sentence, which may not match trivially the current context, forces the QA model to learn more complex relationships than just simple entity matching. The retrieval process may return sentences which do not match the original context. On a random sample, 15/18 retrieved sentences were judged as entirely relevant to the original sentence. This retrieval is already quite good, as we use a high quality ElasticSearch retrieval and use the original context sentence as the query, not just the answer word. While we do not explicitly ensure that the retrieved sentence has
the same meaning, we find that the search results with entity matching gives largely semantically matching sentences. Additionally, we believe the sentences which have loosely related meaning may act as a regularization factor which prevent the
downstream QA model from learning only string matching patterns. Along these lines, \cite{lewis-etal-2019-unsupervised} found that a simple noise function of dropping, masking and permuting words was a strong question generation baseline. We believe that loosely related context sentences can act as a more intuitive noise function, and investigating the role of the semantic match of the
retrieved sentences is an important direction for future work. For the sections which follow, we only show results of retrieved sentences, as the trend of improved performance held across all experiments. 

\begin{table}[t]
\small
\parbox{\columnwidth}{
    \centering
    \begin{tabular}{lll}
    \toprule
    Training procedure                          & EM    & F1    \\
    \midrule
    Cloze-style original              & 17.36 & 25.90 \\    
    Cloze-style retrieved                     
     & \textbf{30.53} & \textbf{39.61} \\
    \bottomrule
    \end{tabular}
    \caption{Effect of original vs retrieved sentences for generic cloze-style question generation.}
    \label{tab:results_cloze_orig_ret}
}
\end{table}
\par
\textbf{Effect of template components:}
We evaluate the effect of individual template components on downstream QA performance.  Results are shown in Table \ref{tab:results_wh_components}. Wh template methods improve largely over the simple cloze templates. ``Wh + B + A + ?" performs best among the template-based methods, as having the Wh word at the beginning most resembles the target SQuAD domain and switching the order of Fragment B and Fragment A may force the model to learn more complex relationships from the question. We additionally test the effect of the wh-component and the question mark added at the end of the sentence. Using the same data as ``Wh + B + A + ?" but removing the wh-component results in a large decrease in performance. We believe that this is because the wh-component signals the type of possible answer entities, which helps narrow down the space of possible answers. Removing the question mark at the end of the template also results in decreased performance, but not as large as removing the wh-component. This may be a result of BERT pretraining which expects certain punctuation based on sentence structure. We note that these questions may not be grammatical, which may have an impact on performance. Improving the question quality makes a difference in performance as seen from the jump from cloze-style questions to template questions. The ablation studies suggest that a combination of question relevance, though matching entities, and question formulation, as described above, determine downstream performance. Balancing those two components is an interesting problem and we leave improving grammaticality and fluency through means such as language model generation for future experiments.
\par
In the last two rows of Table \ref{tab:results_wh_components}, we show the effect of using the wh bi-gram prior on downstream QA training. Using the most-common wh word by grouping named entities into 5 categories according to \cite{lewis-etal-2019-unsupervised} performs very close to the best-performing wh n-gram prior method, while using a single wh-word (what) results in a significant decrease in performance. These results suggest that information about named entity type signaled by the wh-word does provide important information to the model but further information beyond wh-simple does not improve results significantly. 
\par

\begin{table}[t]
\small
\parbox{\columnwidth}{
    \centering
    \begin{tabular}{lll}
    \toprule
    Template data                          & EM    & F1    \\
    \midrule
    Cloze & 30.53   & 39.61 \\
    \midrule
    A + Wh + B + ?          & 45.62 & 55.44 \\
    Wh + A + B + ?          & 44.08 & 53.90 \\
    Wh + B + A + ?          &  \textbf{46.09} & \textbf{56.82}  \\    
    \midrule
    B + A + ?   &  37.57 & 46.41 \\
    Wh + B + A & 44.87 & 54.56  \\
    \midrule
    Wh\_simple + B + A + ? & 45.60 & 56.07 \\
    What + B + A + ? & 10.24  & 17.04 \\
    \bottomrule
    \end{tabular}
    \caption{Effect of order of template, wh word and question mark on downstream QA performance.}
    \label{tab:results_wh_components}
}
\end{table}

\textbf{Effect of filtering by entity matching:}
Besides ensuring that the retrieved sentence and query sentence share the answer entity, we require that at least one additional matching entity appears in both query sentence and entire context. Results are shown in Table \ref{tab:entity_matching}. Auxillary matching leads to improvements over no matching when using template-based data, with best results using matching with both query and context. Matching may filter some sentences whose topic are too far from the original context. We leave further investigation of the effect of retrieved sentence relevance to future work.

\begin{table}[t]
\small
\parbox{\columnwidth}{
    \centering
    \begin{tabular}{lll}
    \toprule
    Matching procedure                          & EM    & F1    \\
    \midrule  
    No matching  & 41.02 & 50.81 \\
    Query matching & 44.76 & 54.87 \\
    Context matching & 44.22 & 55.35 \\
    Query + Context matching & \textbf{46.09} & \textbf{56.82} \\ 
    \bottomrule
    \end{tabular}
    \caption{Effect of query and context matching for retrieved input to question generation module on downstream QA performance.}
    \label{tab:entity_matching}
}
\end{table}

\textbf{Effect of synthetic training dataset size:}
\begin{figure}[t]
\centering
\small
\includegraphics[width=\columnwidth]{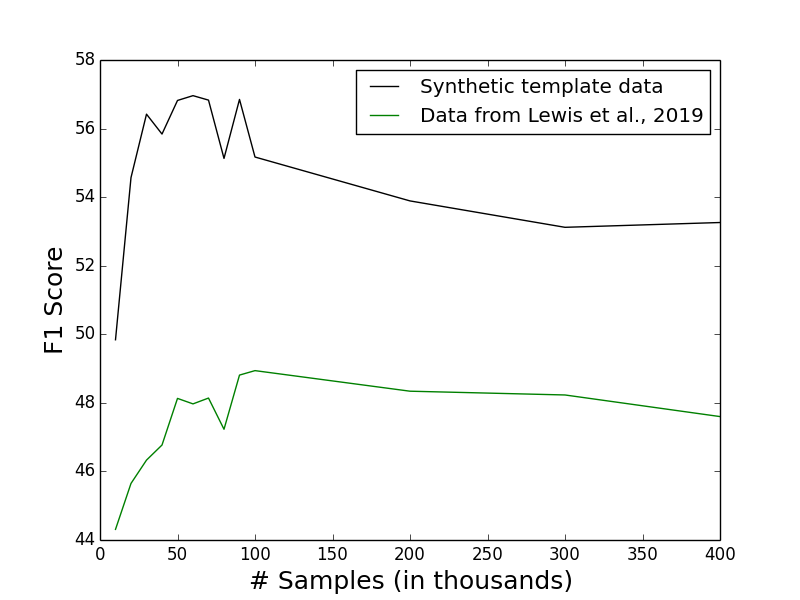}
\caption{A comparison of the effect of the size of synthetic data on downstream QA performance.}
\label{fig:sample_size}
\end{figure} 
Notably, \cite{lewis-etal-2019-unsupervised} make use of approximately 4 million synthetic data points in order to train their model. However, we are able to train a model with better performance in much fewer examples, and show that such a large subset is unnecessary for their released synthetic training data as well. Figure \ref{fig:sample_size} shows the performance from training over random subsets of differing sizes and testing on the SQuAD development data. We sample a random question for each context from the data of \cite{lewis-etal-2019-unsupervised}. Even with as little as 10k datapoints, training from our synthetically generated template-based data with auxiliary matching outperforms the results from ablation studies in \cite{lewis-etal-2019-unsupervised}. Using data from our template-based data consistently outperforms that of \cite{lewis-etal-2019-unsupervised}. Training on either dataset shows similar trends; performance decreases after increasing the number of synthetic examples past 100,000, likely due to a distributional mismatch with the SQuAD data. We chose to use 50,000 examples for our final experiments with other ablation studies as this number gave good performance in initial experiments.
\subsection{Comparison of Best-Performing Models:} 
We compare training on our best template-based data with state-of-the-art in Table \ref{tab:best_results}. SQuAD F1 results reflect results on the hidden SQuAD test set. We report single-model numbers; \newcite{lewis-etal-2019-unsupervised} report an ensemble method achieving 56.40 F1 and a best single model achieving 54.7 F1. We make use of the whole-word-masking version of BERT-large, although using the original BERT-large gives similar performance of 62.69 on the SQuAD dev set. We report numbers on the sample of SQuAD questions which are named entities, which we refer to as SQuAD-NER. The subset corresponding to the SQuAD development dataset has 4,338 samples, and may differ slightly from \cite{lewis-etal-2019-unsupervised} due to differences in NER preprocessing.  We also trained a fully-supervised model on the SQuAD training dataset with varying amounts of data and found our unsupervised performance equals the supervised performance trained on about 3,000 labeled examples. 
\begin{table}[t]
\small
\parbox{\columnwidth}{
    \centering
    \resizebox{\columnwidth}{!}{
    \begin{tabular}{lll}
    \toprule
    Model Choice                          & SQuAD Test F1    & SQuAD NER  F1    \\
    \midrule
    BERT-large (ours)  & \textbf{64.04} & \textbf{77.55} \\
    BERT-large (Lewis et al., 2019) & 56.40 & 64.50 \\
    \bottomrule
    \end{tabular}}
    \caption{A comparison of top results using the BERT-large model.}
    \label{tab:best_results}
}
\end{table}
\par
\section{Conclusion}
In this paper we introduce a retrieval-based approach to unsupervised extractive question answering. A simple template-based approach achieves state-of-the-art results for unsupervised methods on the SQuAD dataset of 64.04 F1, and 77.55 F1 when the answer is a named entity. We analyze the effect of several components in our template-based approaches through ablation studies. We aim to experiment with other datasets and other domains, incorporate our synthetic data in a semi-supervised setting and test the feasibility of our framework in a multi-lingual setting. 

\section{Acknowledgements}
We thank Xiaofei Ma for fruitful discussions on the project. 
\bibliography{acl2020}
\bibliographystyle{acl_natbib}

\appendix
\end{document}